\documentclass[a4paper]{llncs}

\usepackage{amssymb}
\usepackage{graphicx}
\usepackage{algorithm}
\usepackage{algorithmicx}
\usepackage{algpseudocode}
\usepackage{amsmath}
\usepackage{makecell}

\usepackage{url}

\urldef{\mailsa}\path|{sundong.kim}@kaist.ac.kr|    
\newcommand{\keywords}[1]{\par\addvspace\baselineskip
\noindent\keywordname\enspace\ignorespaces#1}

\begin{document}

\mainmatter  

\title{Automatic Knowledge Base Evolution\\ by Learning Instances}

\titlerunning{Automatic Knowledge Base Evolution\\ by Learning Instances}

%
%
\author{Sundong Kim}

\institute{Korea Advanced Institute of Science and Technology, \\291 Daehak-ro, 305-701 Daejeon, Republic of Korea,\\  
\mailsa\\
}

\maketitle

\begin{abstract}
Knowledge base is the way to store structured and unstructured data throughout the web. Since the size of the web is increasing rapidly, there are huge needs to structure the knowledge in a fully automated way. However fully-automated knowledge-base evolution on the Semantic Web is a major challenges, although there are many ontology evolution techniques available. Therefore learning ontology automatically can contribute to the semantic web society significantly. In this paper, we propose full-automated ontology learning algorithm to generate refined knowledge base from incomplete knowledge base and rdf-triples. Our algorithm is data-driven approach which is based on the property of each instance. Ontology class is being elaborated by generalizing frequent property of its instances. By using that developed class information, each instance can find its most relatively matching class. By repeating these two steps, we achieve fully-automated ontology evolution from incomplete basic knowledge base.

\end{abstract}

\section{Introduction}

Knowledge base refinement is the major research area in Semantic web society and many researchers are tackling this problem with diverse methods. Most of these methods are done with semi-automatic way, however the size of the web data is challenging issue. So, fully-automated evolution of knowledge base is a major need in the semantic web society. In addition to that, by looking through the existing knowledge base such as DBpedia[1], you can find many missing information which can be inferred from its internal information. For example, instance `Play station' in DBpedia Korea[10] has multiple properties explaining itself, however its type is still `Thing' which means unclassified. By having schema and property information of both classes and instances, we can map those unclassified instances so that it can be utilized more useful way. 
\\ \indent In this paper, we suggest an algorithm to automatically evolve knowledge base. The algorithm can be categorized as data-driven evolution, since the learning starts from the property information of each instance. By analyzing those information, we can achieve the schema elaboration and suggests better type for each instance. We suggest two main algorithm named property generalization and instance type findings, and these two algorithms works as a mutual way. Consider that we have basic knowledge base which includes ontology information with primitive instances, and new information pile up it as rdf:triple format. After adding the information, two algorithms are executed periodically and give refined knowledge base as an output. 
\\ \indent First algorithm is property generalization algorithm, if there is a famous property among the instances of certain class, we can set the class as a domain type of that property. On the other hand, properties that doesn't represent the instances set lose their domain type information. By adjusting its property, not only each class can be the real representative of its sub-instances, but ontology structure can be refined. 
\\ \indent Second algorithm is type correction of existing instances. Every instances should have its own type, but you can find many mismatch while looking at existing knowledge base especially localized sets. Some instances contain their rdf:type information, but some don't. Some data might have been omitted. Missing value in hierarchy could cause huge information loss while learning. By using our algorithm we can find the best matching between an instance and its relevant classes, and class type is changed if there exists a class which has higher similarity score than instance's former type. To decide the type of instance, we extract the domain information of each property. By analyzing the corresponding domain of each property, we can get the distribution of how each domain affects the instance through property. And by assuming that instance get affected most by the highest-frequency domain, we can designate the type of each instance. 
\\ \indent Our two algorithm work in a complementary way. By generalizing the property of the instance into the property of the class, our ontology is getting richer, hence each instance can use more information to find their new class. Unclassified instance can also be classified throughout the cycle. And more relevant instances are tagged into each class so that the next generalization procedure operates more accurately. Figure 1 illustrate briefly how our whole evolution logic works.

\begin{center}
\includegraphics[scale=0.35]{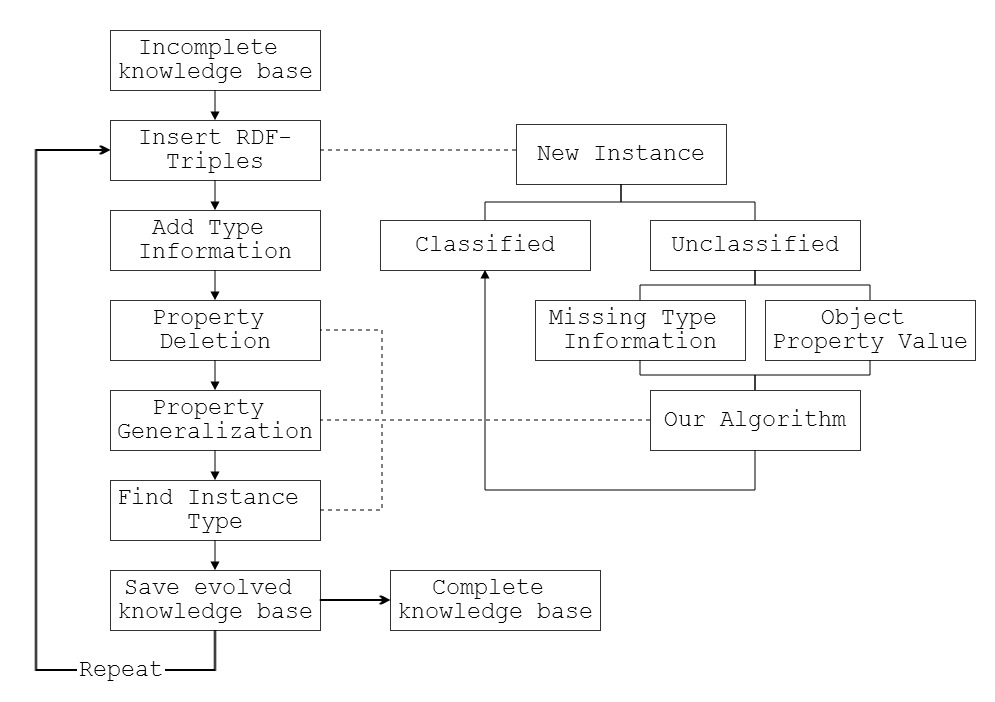}
\begin{small}
\textbf{Fig. 1.} The cycle of fully-automatic ontology evolution 
\end{small}
\begin{figure}
\centering
\label{fig:example}
\end{figure}
\end{center}

\section{Related Works}

Traditional ontology evolution method can be categorized as three different types[2]: data-driven evolution, structure-driven evolution, usage-driven evolution. Structure-driven evolution is the evolution method by analyzing the structure of ontology. For example, by using structure-driven evolution, we can make more understandable hierarchy. If certain class has many subclasses, we can make one more sub-level for controlling the number of subclasses. Data-driven approach is mostly based on instance learning. For example, if all instances of class A are the member of class B, class A becomes the subclass of class B. Stojanovic especially, emphasized data-driven approach should be done by direct instances. External data sources are used by various research groups to complete data-driven approach. Evolva[3] used RSS feeds, terms list to initiate ontology evolution. And many ontology learning tools such as Text2Onto[4], SPRAT[5] shows ontology changes from text corpus. For extending the coverage of the localized version of DBpedia, Airpedia [6, 7] did mapping between Wikipedia infobox to 14 different languages. And [8] proposed semi-automatic schemata construction by extracting axiom patterns in existing knowledge base and converted into SPARQL based pattern detection algorithms which allows to refine knowledge base. 
 And User-driven evolution use query and user-log information. Grouping criteria for knowledge base can be the frequency of user access of the information. To do so, searching time can be reduced in the way that higher-accessed group has priority. Web usage mining[9] is the famous method of usage-driven evolution.

\section{Evolution Algorithm}
For automatic schema evolution, property generalization algorithm and type correction algorithm alternatively operates. We suggest three different instance type finding methods, and probability-based property generalization method.

\subsection{Finding Instance Type}
Naturally, each DBpedia instance can be categorized as a certain type, and each type has property set that describes it. So each instances has property information related to its type. But some instance have various property that can't be categorized just in a type. Or type data could be missed while building ontology. For example, DBpedia instance Arnold\textunderscore Schwarzenegger[11] is categorized as rdf:type 'Agent', 'Person' and 'Officeholder'. To clarify it, DBpedia just simply put Arnold's matching Yago[12] type information on the list though it can't be exactly match to another DBpedia type that describes Arnold\textunderscore Schwarzenegger better. As you know, Arnold\textunderscore Schwarzenegger can be categorized as politician, bodybuilder and actor. In this case, we can't apply traditional data-driven and structure-driven evolution algorithm. So finding instance type using its internal information is extremely important for further analysis. By looking through the instance, property information is valuable to perform knowledge-base evolution. In the schema level, each property has several domain type. By looking through DBpedia ontology[13], DBpedia property 'birth date' has domain 'person', but 'goals in league' has more specific domain 'soccer player'. We assume that property of each instance indirectly explain hidden type of instances. So we suggest three different ways of finding instance type through analyzing instance's property.   

\subsubsection{Naive Frequency Counting}

In naive frequency counting algorithm, we analyze the property - domain information of each instance and select the most relevant domain as an instance type.
Table 1 shows how this algorithm works for sample instance 'Dae-jung Kim'[14]. The left table shows the list of properties that the sample instance has, and extracted domain information for each property. After listing, we simply count the frequency of domain appeared. Right table shows the list of domain in a descending order. As expected, we can find the correct domain type 'President' for the former Korean president 'Dae-jung Kim'. Also, we can utilize the relevant class information extracted from the result.

\begin{center}
\small
\begin{tabular}{| l | r | l | l |}
\hline 
Property Name & \# Domains & Domains \\ 
\hline 
foaf:name & 101 & \textbf{President}, Person \\ 
\hline 
picture: & 61 & Artist, Person \\ 
\hline 
country & 34 & \textbf{President}, Officeholder  \\ 
\hline 
birthPlace & 31 & \textbf{President}, Monarch \\ 
\hline 
diedIn & 29 & \textbf{President}, Monarch \\ 
\hline 
birthDate & 28 & \textbf{President}, Person \\ 
\hline 
... & ... & ... \\ 
\hline 
InaugurationDay & 2 & \textbf{President}, Officeholder  \\ 
\hline 
vicePresident  & 1 & \textbf{President} \\ 
\hline 
\end{tabular} 
\quad
\begin{tabular}{| l | r |}
\hline 
Domain & Count \\ 
\hline 
\textbf{President} & 25 \\ 
\hline 
OfficeHolder & 15  \\ 
\hline 
Politician & 14  \\ 
\hline 
Monarch & 14 \\ 
\hline 
Officer & 9 \\ 
\hline 
... & ...  \\ 
\hline 
Actor & 1 \\ 
\hline 
Model & 1 \\ 
\hline 
\end{tabular} 

\medskip
\begin{small}
\textbf{Table 1.} Finding domain type of DBpedia instance 'Dae-jung Kim'   
\end{small}
\end{center}

\begin{algorithm}
\caption{Naive Frequency Counting}\label{nfc}
\begin{algorithmic}[1]
\Procedure{Naive Frequency Counting}{}\Comment{Find Instance Type}
   \ForAll{direct instances of class} 
   \State Instantiate Map\textless Property, Frequency\textgreater ~m
      \ForAll{rdf-Properties of instance} 
         \ForAll{domain of property}
         \State Put property-domain pair frequency into m
         \EndFor
      \EndFor
      
      \ForAll{property-domain pair in m}
      	 \State Aggregate frequency for each domain
      	 \If {frequency is the highest}
      	    \State change instance domain into class
      	 \EndIf 
      \EndFor 
   \EndFor
  
\EndProcedure

\end{algorithmic}
\end{algorithm}

\subsubsection{Cosine Similarity}
By naive frequency counting, we can find the most related class but it can't directly be accepted as a similarity score. To get the rigorous calculation, we can formulate two matrices and get Type-Instance similarity score. First, we formulate two matrices, type-property matrix and instance-property matrix. Value of those adjacency matrix is 1 if we consider Naive Frequency Counting case. 
However, by normalizing row vectors of two matrices, we can give an relative weight that the type domain which has a lot of property can't get advantages. This method exactly represents Cosine similarity between instance and type regarding to property information, and the score is bounded by 0 to 1. On the example, property set of Instance A is exactly same as property set of Type 1. In that case, we can get similarity score 1 between instance A and type 1.

\subsubsection{Cosine similarity with TF-IDF}
TF-IDF can give a weight to the term which only exists in specific set of documents. In our example, it becomes PF-IDF(Property Frequency-Inverse Domain Frequency). By considering inverse domain frequency, we can give weight to the property which has a few domain. So, type-property matrix can be recalculated using TF-IDF measure. Although TF-IDF measure is effective to measuring weight, instance-property matrix doesn't need to be recalculated, since our intuition is that although common property are shared by instances, importance of that property will not be underestimated. Only if that property is shared among various types, we can judge that the property affects minimally while choosing its type. So, final score will be calculated by cosine similarity between PF-IDF type-property matrix and instance-property matrix.
For example, an uncategorized instance have specific property 'goals in league', we can say that the instance has a probability of being categorized as 'soccer player'. On the other hand, if instance has general property like 'birth date', it doesn't affects much for classifying its type. Figure 2 shows the effect of applying TF-IDF into the sample instance 'Daejung Kim'. Left figure shows the property set without adjusting and right figure shows the property set after applying inverse domain frequency. As a result, common property such as 'description', 'bornIn' are disappeared after adjusting its weight.

\begin{center}
\includegraphics[scale=0.5]{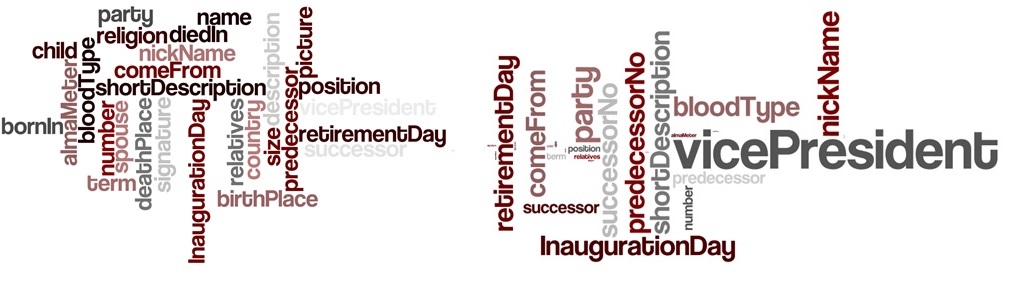}
\begin{small}
\textbf{Fig. 2.} Property set of instance 'Daejung-Kim' before and after TF-IDF
\end{small}
\begin{figure}
\centering
\label{fig:example}
\end{figure}
\end{center}    
\noindent

\smallskip

\subsection{Property Generalization}
To infer instance type correctly, we require solid ontology that gives huge property information. Property generalization is the way of data driven approach, by reinforcing ontology by generalizing and deleting property through instance-based learning. In detail, famous property shared by most of the instances in certain type gets domain type information after generalized. Also, property can lose it's domain information if greater part of the instances of that class doesn't have it as a characteristic. Property generalization operates in two different conditions. First, when new triples are added to the current knowledge base and detailed information of the instance gets bigger, we take property generalization to make ontology updated for new information. Second, after adjusting instance type we have to consider the effects to knowledge base. For example, new instances are added, and existing instances are deleted from certain type and we want to modify the knowledge based on those minor changes, so that ontology can reflect the recent updates in knowledge base. To adjusting the ontology, we suggest probability generalization ratio, \begin{center}
$P = \dfrac{1}{1+\log_{10}N}, $  \  N : number of instances for a type     
\end{center}
if instances more than probability P have certain property, we can generalize the property into ontology which means property can get a new domain type. Property deletion activates on the opposite way by deleting domain type if it is not famous throughout the instances. Property generalization and deletion algorithm activate from the leaf nodes so that ontology refinement can occur sequentially. In our full view, ontology is refined through property analysis and by using evolved ontology we can adjust instance type information. By repeating this procedure with new instance triples, knowledge base can evolve in fully automated way.
\begin{algorithm}
\caption{Property Generalization}\label{nfc}
\begin{algorithmic}[1]
\Procedure{Property Generalization}{}\Comment{Enrich Ontology}
   \State Instantiate Map\textless Property, Frequency\textgreater ~m
 
   \ForAll{direct instances of class} 
      \ForAll{rdf-Properties of instance} 
         \State Put property and its frequency into m
      \EndFor
   \EndFor
      
   \ForAll{Property in m}
   	 \If {metThreshold() is true}
   	    \State Add domain class to property
   	 \EndIf 
   \EndFor 
\EndProcedure

\end{algorithmic}
\end{algorithm}

\begin{algorithm}
\caption{Automatic Knowledge-Base Evolving System}\label{nfc}
\begin{algorithmic}[1]
\Procedure{Automatic Knowledge-Base Evolving System}{}
   \State Load existing Knowledge-Base
   \While{There exists instance triples to add}
      \State Add Instance triple sets
      \ForAll{classes from the leaf}
         \State \textsc{Property Generalization()}
         \State \textsc{Property Deletion()}
      \EndFor
      \ForAll{classes from the leaf}
         \State \textsc{Find Instance Type()}
      \EndFor      
   \EndWhile  
   \State Save evolved Knowledge-Base
\EndProcedure

\end{algorithmic}
\end{algorithm}

\section{Experiment Setup}
To validate our algorithm, we setuped DBpedia ontology by using Prot{\'e}g{\'e}[15] environment, and added instances incrementally to check how many unclassified instances are classified by the effect of our algorithm. English-Korean mapping information[16] is used while implementing the system.

\subsection{Find the type of unclassified instances} 
At the first experiment, we added available instance type information from the first, and add 50,000 lines of new triples each time and update the knowledge-base by using our algorithm. Since not every instance has rdf:type information they remained as unclassified after they added to the knowledge base. For example, some instance missed their instance type and some are being generated by the value of object property. Latter instances have no property if there aren't exist before, it will takes time to get property filled. Finally, by using the property characteristic of the ontology, those unclassified instances can find its own type.     
  
\section{Results and Discussion}
We checked every cycle and how many instances get type value extracted throughout the evolution cycle. Figure 3 illustrates that 82.4\% of unclassified instances which has properties find its own domain type. We can interpret that the area between green and black line is the number of instances newly classified by our evolution algorithm. The area between green and red line is yet unclassified, since the property doesn't have much information, and some instances between red and blue line have no property since they are newly generated from object property value.
  
\begin{center}
\includegraphics[scale=0.5]{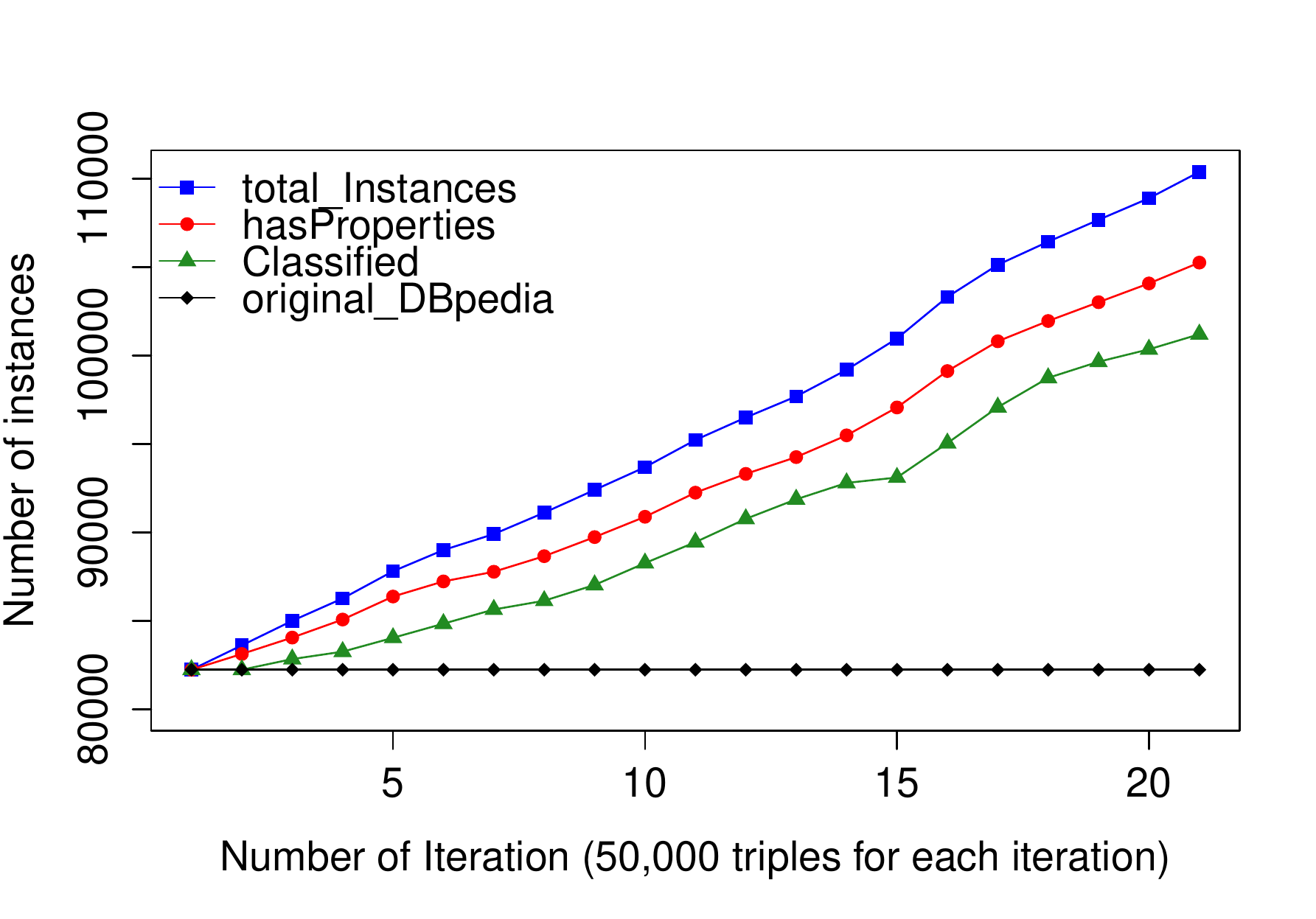}

\begin{small}
\textbf{Fig. 3.} Increasing classified instances in DBpedia Korea throughout the iteration
\end{small}
\begin{figure}
\centering
\label{fig:example}
\end{figure}
\end{center}    
\noindent
\smallskip

Figure 4 illustrates the number of property having domain increases throughout the iteration. The ratio also increases from 76.9\% to 91\%. For each iteration, ontology information overall increases through property generalization algorithm. 

\begin{center}
\includegraphics[scale=0.5]{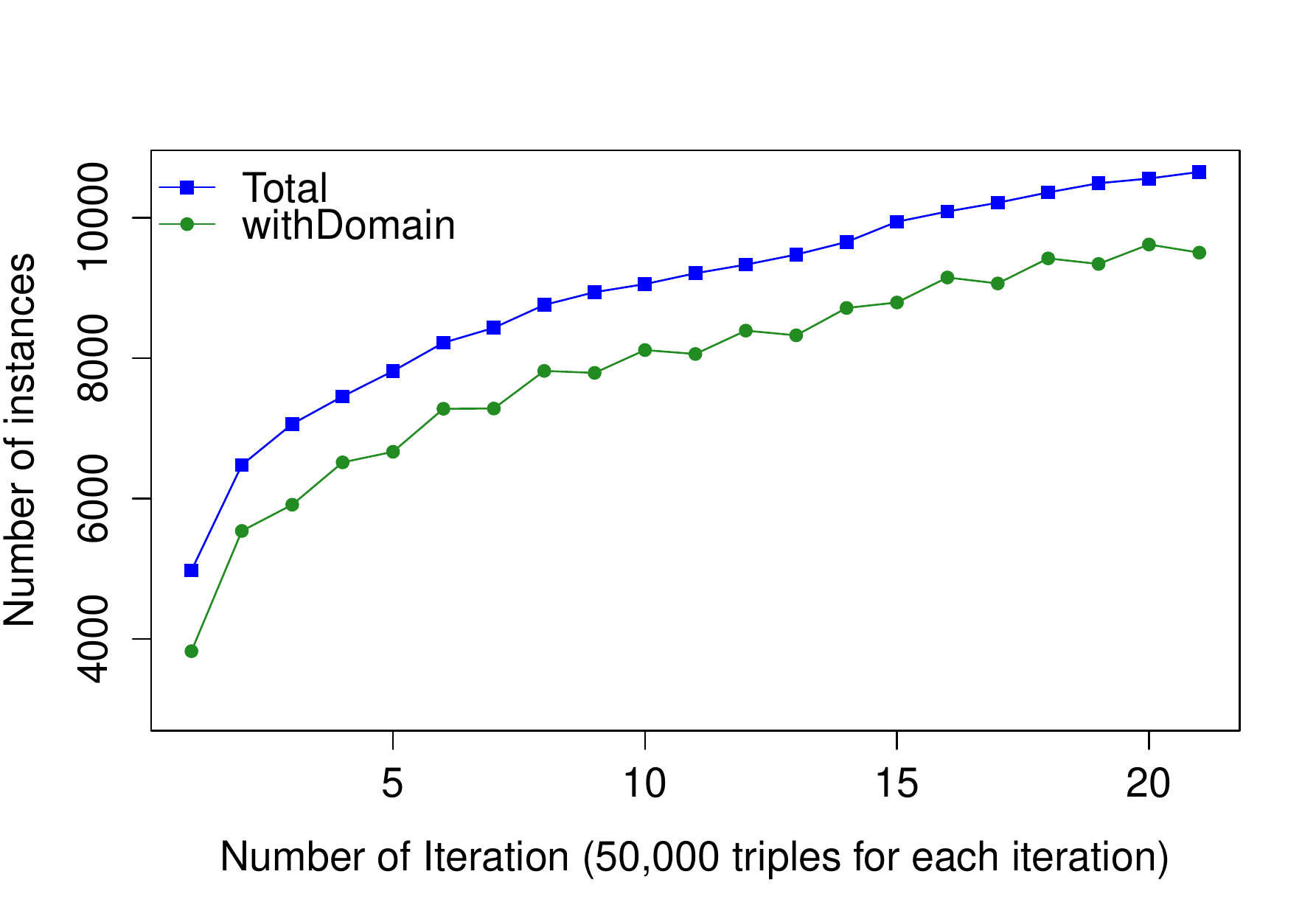}

\begin{small}
\textbf{Fig. 4.} Increasing classified property in DBpedia Korea throughout the iteration
\end{small}
\begin{figure}
\centering
\label{fig:example}
\end{figure}
\end{center}    
\noindent

\smallskip

\section{Conclusions and Future Work}
In this paper, we have introduced a new fully-automated knowledge-base evolution methods, which operates in two steps - property generalization and instance type finding. Using this method, we can generate evolved knowledge-base incrementally from incomplete ontology  According to this method, property plays an important role to refine ontology. 
\\ \indent The major advantage of this approach is that knowledge base can easily adapted to the new information based on probabilistic model. Not only suggesting link between new information and existing ontology, we can refine the ontology and instance information. This allows our methods as a start point to generate time-evolving knowledge-base. 
\\ \indent We validated our algorithm through DBpedia dataset, and proved the effectiveness of our algorithm. One weakness of the method is that the ontology itself is heavily depends on the information we get. So validation of the rdf:triple is needed before getting into the system.
\\ \indent Further work includes the mapping new class into ontology on the right position, as well as validating our ontology if the input triple come from the text corpus which doesn't related to any pre-existing knowledge base.

\subsubsection*{Acknowledgments.}

 This work was supported by the IT R\&D program of MSIP/KEIT. 
[10044494, WiseKB: Big data based self-evolving knowledge base and reasoning platform]

\label{References}

\end{document}